# Feature Engineering vs. Deep Learning for Paper Section Identification: Toward Applications in Chinese Medical Literature

Sijia Zhou; Xin Li

Department of Information Systems, City University of Hong Kong, Hong Kong;

City University of Hong Kong Shenzhen Research Institute, Shenzhen

sjzhou3-c@my.cityu.edu.hk; Xin.Li.PhD@Gmail.com

## Abstract

Section identification is an important task for library science, especially knowledge management. Identifying the sections of a paper would help filter noise in entity and relation extraction. In this research, we studied the paper section identification problem in the context of Chinese medical literature analysis, where the subjects, methods, and results are more valuable from a physician's perspective. Based on previous studies on English literature section identification, we experiment with the effective features on classic machine learning algorithms to tackle the problem. It is found that Conditional Random Fields, which consider sentence interdependency, is more effective in combining different feature sets, such as bag-of-words, part-of-speech, and headings, for Chinese literature section identification. Moreover, we find that classic machine learning algorithms are more effective than generic deep learning models for this problem. Based on these observations, we design a novel deep learning model, the Structural Bidirectional Long Short-Term Memory (SLSTM) model, which models word and sentence interdependency together with the contextual information. Experiments on a human-curated asthma literature dataset show that our approach outperforms the traditional machine learning methods and other deep learning methods and achieves close to 90% precision and recall in the task. The model shows good potential for use in other text mining tasks. The research has significant methodological and practical implications.

# 1 Introduction

The big data era has led to a boom of unstructured data, including textual data. To make use of textual data, it is often necessary to structuralize the free text to support modeling and analysis. In library science and literature analysis, researchers are interested in extracting information and knowledge from books and scientific publications. Publications generally/tend to have an inherent logic structure, and the importance level varies for different sections.Thus, section identification on free text (TAKAHIKO Ito, Shimbo, & Amasaki, 2004) could significantly impact search efficiency for users.

In the medical domain, scientific literature analysis is helpful to researchers who want to digest medical literature. Due to the large and ever-increasing volume of medical literature, automatic information and relation extraction (H. Zhang, Boons, & Batista-navarro, 2019) are needed, especially specific entity extraction (Shi & Bei, 2019). However, automatic analysis tools for literature written in Chinese are less developed as compared with those for Western literature (often in English). While there is an urgent need to exploit Chinese medical resources, the lack of tools hinders the reuse of knowledge documented in Chinese medical literature.

This study focuses on improving the methods for section identification, which can benefit Chinese medical literature analysis. In this context, the experiment and results sections, which can be considered as clinical guidelines in evidence-based medicine (EBM), are important for medical researchers. Other sections, such as literature review, are of lower value (X. Zhou, Peng, & Liu, 2010). Differentiating the sections can help filter the most relevant parts for further analysis. Some previous efforts built machine learning models to identify sections of English medical documents (Chung, 2009; Hirohata, Okazaki, Ananiadou, & Ishizuka, 2008; S. N. Kim, Martinez, & Cavedon, 2010; Lui, 2012). Due to natural language differences, their extracted features cannot be directly applied onto Chinese medical literature. Moreover, Chinese medical literature is often less structured than English literature, which makes the problem more difficult to tackle.

Noticing the gap in previous research, this paper explores two types of efforts to tackle the section identification problem. First, we take a feature engineering approach and investigate the most effective features that can be used for Chinese medical literature section identification. Second, we take a deep learning approach and develop a novel Structural BLSTM (SLSTM) model that considers the dependencies between sentences for section identification. Through experiments on an asthma treatment dataset, we identified the set of effective features that work best with the Conditional Random Fields (CRF) model. Moreover, our proposed SLSTM model significantly outperforms all classic methods (including the CRF model) and the state-of-the-art deep learning models.

The contribution of the paper is three-fold: 1) The paper provides a practical guideline on the effective features to use in the feature engineering paradigm for Chinese medical literature section identification. We also show that this traditional approach performs very well in this task as compared with deep learning methods. 2) The paper shows that it is important to consider the dependencies between sentences for section identification in both feature engineering and deep learning approaches. The deep learning model outperforms traditional models when this feature is considered. Although the dependencies of sentences are considered in previous feature engineering studies for section identification, they were not studied in a deep learning context. Moreover, most previous studies taken this approach are based on short texts, such as abstracts. Our studies increase the generalizability of this finding to full text of Chinese traditional medicine documents. 3) This paper proposes a SLSTM deep learning model that is very effective for section identification. While we believe there exist other more effective models to be explored in future research, our study shows the potential of this particular model, which may be applicable to address other generic sentence classification problems.

This paper is organized as follows. In Section 2, we introduce the related work. In Section 3, we describe the problem setting. In Section 4, we introduce our feature engineering and deep learning approaches. Then we describe the experiments in Section 5. Results are in Section 6. Finally, we conclude this research in Section 7.

## 2 Related Work

### 2.1 Existing Section Identification Studies

Section identification is a text mining task in scientific literature analysis that seeks to identify the roles of different sentences in a paper's logic structure (TAKAHIKO Ito et al., 2004). This problem is more distinct for scientific literature, which often contains a standard logic. In general, section identification is considered as a sentence classification problem, where each sentence is assigned a section label.

In order to review existing studies, we develop a taxonomy as shown in Table 1. From a research subject perspective, we focus on the language, document type, and domain of the articles to conduct section identification work. In general, most previous studies focus on English materials and often use the abstract of medical literature as training data, since many journals require abstracts to have standardized formats, such as subjects, methods, results, etc. There are a few studies on legal documents (Hachey & Grover, 2005), literary documents like biographies (L. Zhou, Ticrea, & Hovy, 2005), and Italian radiology reports (Esuli, Marcheggiani, & Sebastiani, 2013). From a modeling perspective, we focus on machine learning algorithms and the features they can use. In particular, the information may be textual features from paper content and other types of information. As shown in the table, we further classify textual features to linguistic features, structural features, and semantic features. Linguistic features represent the features on linguistic patterns of sentences, such as bag-of-words (BOW), n-unigram, and part-of-speech (POS). Structural features describe the locational and structural characteristics of sentences, such as the length, section heading information, and location of the sentence in the paper. Semantic features show the semantic meaning of the terms in papers. Previous studies also considered features such as citations and inter-sentence dependencies. Citation features represent references embedded in the papers. Sentence dependency considers if adjacent sentences can be employed for labeling. When modeling sentence dependency, features of nearby sentences can be included in focal sentences' classification. It is also possible to capture sentence dependencies using models such as CRF and Hidden Markov Models (HMM).

Table 1. A Taxonomy to Review Section Identification Studies

| *Category* | *Subcategory* | *Explanation* |
|---|---|---|
| Language | | The language of the document |
| Document Type | | Abstract or full text |
| Domain | | Academic field of the documents |
| Textual Features | Linguistic feature | Bag-of-words; Part-of-speech; Lexicon; N-gram |
| | Structural feature | Heading; Sentence location; Sentence length; etc. |
| | Semantic feature | Synonyms; Hypernyms |
| Other Information | Citation | References embedded in papers |
| | Sentence dependency | Interdependency between sentences |
| Model | | Machine learning algorithms |

Based on the taxonomy, we review existing studies in Table 2. From an application domain perspective, most studies focus on English literature, particularly abstracts. There is an obvious gap in applications to Chinese literature, especially full texts of papers. The table also shows the use of features and models, which can guide this study.

From the textual feature perspective, linguistic features are most frequently used, although the actual features vary. For example, Xu et al. (2006) adopted different types of bag-of-words (BOW) features, such as stemmed words and BOW excluding stop words. Hassanzadeh et al. (2014) employed passive verbs and active verbs as linguistic features. Nam et al. (2016) employed the domain lexicon as features, including some typical verb phrases and noun phrases. Malmasi et al. (2015) explored the clustering of linguistic features as a new features task that could improve section identification performance.

Structural features are the second most popular, since papers often have a standard structure. For some medical abstracts, the length of each section is also predetermined, which eases the problem. McKnight and Srinivasan (2003) employed sentence location (i.e., first sentence, middle sentences, last sentence) to identify sections from medical abstracts. Sentence length can also be considered, especially for some professional materials (Hachey & Grover, 2005). Headings are another important structural feature. Lui (2012) considered the combination of linguistic features (BOW, POS, and bigrams of POS tagged lemmas) and structural features (sentence length, headings, and abstract/article length) to achieve better performance.

Some researchers explored semantic smoothing methods to generate semantic features such as synonyms and hypernyms for the task. For example, Kim et al. (2010) employed semantic features to deal with proper nouns. They referred to the Unified Medical Language System (UMLS) biomedical ontologies to identify semantic relationships between terms. Then they got Concept Unique Identifiers (CUIs) that mapped different text into the ontological concepts, which are one of the semantic features in existing research.

Besides textual features, other information can be employed. The most frequently considered information is sentence dependency. To use sentence dependency, some studies considered features in nearby sentences, while others considered the information from the model structure. For example, Shimbo et al. (2003) used features extracted from the previous and next sentence with SVM to classify sections that would be of interest to users. This study contributed to the building of a search system for the MEDLINE database. From the model structure perspective, the Hidden Markov Models (HMM) can consider the label of the previous sentence to classify the current sentence. It is thus being employed in the section identification problem (Lin et al., 2006). Later, Hirohata et al. (2008) examined the use of previous and following sentences in different window sizes under a CRF framework, due to the fact that CRF can incorporate more complicated sentence dependencies than HMM. Chung (2009) evaluated the combination of n-gram features, POS, sentence location, and windowed features, identifying the importance of considering the sentences dependency for section identification.

Citation features can also be adopted in section identification, if the citation information is important for the research (Angrosh, Cranefield, & Stanger, 2010; Hachey & Grover, 2005). For example, Angrosh et al. (2010) adopted two Boolean features to represent whether the current sentence or its previous sentence has a citation. Hachey and Grover (2005) chose a citation feature called quotation to show the percentage of sentence tokens inside an inline quote.

Table 2 clearly shows the evolution of machine learning algorithms. In the early stage, people mostly used classic algorithms, such as SVM (Takahiko Ito, Shimbo, Yamasaki, & Matsumoto, 2004; Yamamoto & Takagi, 2005). Later HMM was popular in section identification studies (Lin et al., 2006; Wu et al., 2006; Xu et al., 2006). More recently, CRF has been the dominant method (Angrosh et al., 2010; Chung & Coiera, 2007; Esuli et al., 2013; S. N. Kim et al., 2010). We believe the reason for the change of methods is due to the models' abilities to capture sentence dependency.  In the past two years, deep learning models have been used in this problem. For example, Agibetov et al. (2018) used a deep learning technique, word embedding, to process textual features before feeding them into a neutral network to tackle the problem. However, existing studies that use deep learning for section identification are still limited.

Table 2. Summary of Previous Research on Scientific Literature Section Identification

| Studies | Lang | Document Type | Domain | Textual Feature | | | Other Infor | | Model |
|---|---|---|---|---|---|---|---|---|---|
| | | | | L | S | Se | C | SD | |
| (Agibetov et al., 2018) | English | Abstract | Medical | | | | | | Word embedding + Neutral Network |
| (Dernoncourt, L, & Szolovits, 2017) | English | Abstract | Medical | | | | | | BLSTM |
| (Nam et al., 2016) | English | Abstract | Medical | √ | √ | | | | SVM |
| (Malmasi et al., 2015) | English | Abstract | Medical | √ | | | | | SVM |
| (Hassanzadeh et al., 2014) | English | Abstract | Medical | √ | √ | | | √ | CRF; SVM; Naïve Bayes; Logistic regression |
| (Esuli et al., 2013) | Italian | Full paper | Medical | √ | √ | | | √ | CRF |
| (Lui, 2012) | English | Abstract | Medical | √ | √ | | | | Logistic regression; Naïve Bayes; SVM |
| (Angrosh et al., 2010) | English | Full paper | Computer | √ | | | √ | √ | CRF |
| (S. N. Kim et al., 2010) | English | Abstract | Medical | √ | √ | √ | | √ | CRF |
| (Chung, 2009) | English | Abstract | Medical | √ | √ | | | √ | CRF |
| (Hirohata et al., 2008) | English | Abstract | Medical | √ | √ | | | √ | SVM; CRF |
| (Chung & Coiera, 2007) | English | Abstract | Medical | √ | | | | √ | SVM; CRF |
| (Ruch et al., 2006) | English | Abstract | Medical | √ | √ | | | | Bayesian Model |
| (Lin et al., 2006) | English | Abstract | Medical | √ | | | | √ | HMM |
| (Xu et al., 2006) | English | Abstract | Medical | √ | | | | √ | HMM; Naïve-Bayes; Maximum Entropy; Decision Tree |
| (Wu et al., 2006) | English | Abstract | Medical | √ | | | | √ | HMM |
| (L. Zhou et al., 2005) | English | Full paper | Literature | √ | | | | | Naive Bayes; SVM; Decision Tree |
| (Hachey & Grover, 2005) | English | Full paper | Law | √ | √ | | √ | √ | CRF |
| (Yamamoto & Takagi, 2005) | English | Abstract | Medical | √ | √ | | | | SVM |
| (Ito et al., 2004) | English | Abstract | Medical | √ | √ | | | √ | SVM |
| (Shimbo et al., 2003) | English | Abstract | Medical | √ | √ | | | √ | SVM |
| (McKnight & Srinivasan, 2003) | English | Abstract | Medical | √ | √ | | | | SVM; Linear models |

(L=Linguistic features; S=Structural features; Se=Semantic features; C=Citation; SD= Sentence dependency)

## 2.2 Deep Learning for Text Classification

Although deep learning is not widely used in section identification, it has been applied in many different domains including text classification. The classic deep learning model is Convolutional Neural Networks (CNN), which has been studied in text mining. For example, Kim (2014) employed CNN for sentence-level classification, including both polarity and topic classification. Yin and Schütze (2015) considered multichannel feature maps of sentences under a CNN framework for sentence classification and reduced the impact of unknown words. Zhang et al. (2015) used character-level CNN for text classification, which achieved competitive results with the state-of-the-art methods while avoiding the need for word segmentation.

For text mining, CNN does not effectively capture the sequence of words and sentences. Recurrent Neural Networks (RNN) can capture such sequential information to an extent. Thus, researchers, such as Lai et al. (2015), also employed RNN for text mining. From the perspective of modeling dependencies in text, the Long Short-Term Memory (LSTM) model is more effective. The main idea of LSTM is to introduce an adaptive gating mechanism that determines the degree to which the previous state is maintained and then memorizes the extracted features for the current input. The bidirectional LSTM (BLSTM) model further introduces a second hidden layer to capture the dependencies among words in text in two directions. Since BLSTM can consider information from both the previous and next words, it has been successful in recent deep learning research and is considered the state-of-the-art algorithm in deep learning text mining (P. Zhou et al., 2016). BLSTM has been shown to be more effective than the traditional feature-based approach in tasks such as relation classification (P. Zhou et al., 2016) and opening mining (Nguyen & Nguyen, 2018). Dernoncourt and Szolovits (2017) used BLSTM on biomedical texts to address the section identification problem.

## 2.3 Research Gap and Research Objectives

The literature highlights the gap in existing studies and suggests the objectives we need to achieve in this study.

First, there are limited studies on paper section identification in Chinese literature, especially on Chinese medical literature. It is necessary to explore if the features used in English paper section identification, such as linguistic features, structural features, and sentence dependency, will be able to address section identification in Chinese papers.

Second, most existing methods (in English paper section identification) are classic machine learning algorithms based on feature engineering. Previous studies show good potential for using deep learning in text classification problems. However, it is not clear whether the model can effectively deal with the section identification problem. It would be beneficial to explore and compare feature engineering and deep learning approaches in section identification.

Third, previous studies on feature engineering methods generally focus on the paper structure and interdependencies between sentences as informative features for section classification. It would be beneficial to explore if such design rationales can be implemented in a deep learning approach and whether they can improve section identification in Chinese literature.

# 3 Problem Setup

## 3.1 The Section Identification Task for Chinese Medical Literature

Figure 1. Two Chinese Medical Literature Example Papers

Figure 1 illustrates two typical Chinese medical papers. As we can see, after stating the purpose of the study, Chinese medical papers often explain details of the experiments, including the subjects, the treatment, and the effect of treatment. This is the most important information that has the potential to benefit clinicians or researchers if extracted (El-Gayar & Timsina, 2014). A paper may also discuss experiment mechanisms and theoretical reflections. In this study, we aim to identify the sentences that belong to the following three labels:

- Subject: The humans or objects that comprise the study sample.
- Method: The treatment and medicine that the study used to treat the disease.
- Result: The effect of the treatment, usually including the cure rate and improvement rate.

**3.2 Section Identification as Sentence Classification**

Following previous research (Chung, 2009; Hirohata et al., 2008; Takahiko Ito et al., 2004; S. N. Kim et al., 2010; Lui, 2012; McKnight & Srinivasan, 2003; Yamamoto & Takagi, 2005), we address the section identification problem as a sentence-labeling problem and build classifiers to classify all sentences in each paper. Noting that the three classification labels cannot cover all sentences in a paper, we arbitrarily define two supportive labels, "Pre" and "After", depending whether a sentence is before the first sentence of "Subject/Method/Result", or after the last sentence of "Subject/Method/Result." If the sentence is not belonging to these five categories, it is naturally put into an "other" category by a classifier. In this setup, a regular user would be more interested in the classification results of Subject, Method, and Result. Pre and After can be considered as hidden labels to eases the framing of the problem to a multi-class sentence classification problem."

# 4 Method

Figure 2 depicts the general process for our task of addressing section identification as sentence classification. To facilitate this task, we first conduct sentence segmentation based on Chinese punctuation such as "。", "？", and "!" If a paragraph contains only a single line without punctuation, we also consider it as a sentence.

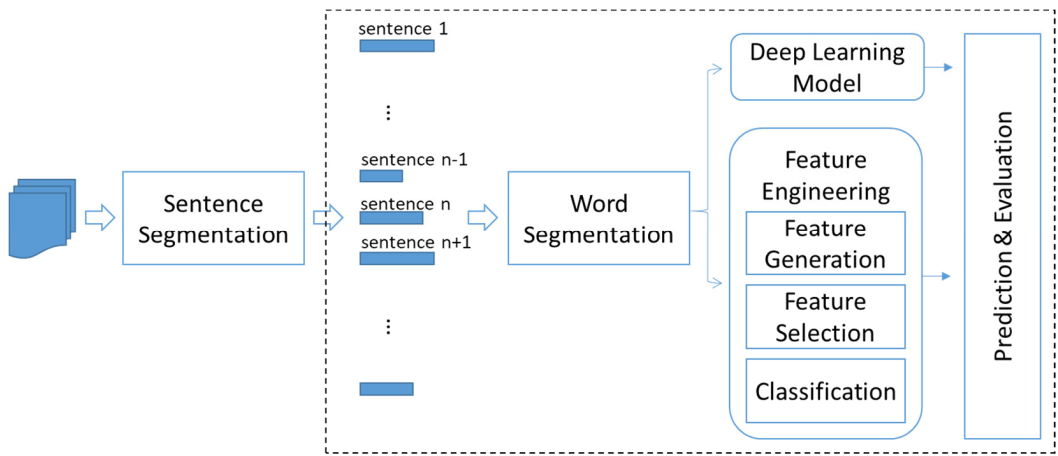

Figure 2. The General Process for the Section Identification Task

Next, we develop machine learning models to classify each sentence of each paper. Since our contents are in Chinese, we first conduct word segmentation using Jieba (https://github.com/fxsjy/jieba/). To facilitate the word segmentation task, we develop a dictionary using the traditional medical terms and major lexicons available online. The dictionary contains about 10 million Chinese words. In this study, we take two approaches to build the sentence classifier: the feature engineering approach and the deep learning approach, the freer state-of-art methodology. For the feature engineering approach, we generate features from the sentences and conduct feature selection to choose the most relevant features to build the machine learning classifier. For the deep learning approach, we design a deep learning model that performs the classification task directly without feature extraction. The models are evaluated on their prediction performances with human-coded Chinese medical literature.

**4.1 Feature Engineering Approach**

*4.1.1 Feature Generation*

Feature generation is the most important step in the feature engineering approach. Based on our literature review, we extracted the linguistic features and structural features from individual medical papers as shown in Table 3. In this study, we do not have citation information or Chinese medical ontologies that can support the extraction of citation information and semantic features.

Table 3. Textual Features Extracted from Individual Documents

| *Feature Sets* | *Features* | *# of Generated Features* | *# of Selected Features* |
|---|---|---|---|
| Linguistic features | Bag-of-words (BOW) | 14,172 | 68 |
| | Part-of-speech (POS) | 15,889 | 87 |
| | Latent Dirichlet Allocation | 40 | 0 |
| | Doc2Vector | 40 | 0 |
| Structural features | Heading | 274 | 15 |
| | Sentence Position | 1 | 0 |
| | Sentence Length | 2 | 0 |

First, we use linguistic features. Based on their performance in previous studies, we choose four types of linguistic features: bag-of-words, part-of-speech, Latent Dirichlet Allocation (LDA), and Doc2Vector.

- *Bag-of-words (BOW)* is a widely used representation for documents (Chung, 2009; S. N. Kim et al., 2010; Lui, 2012), which considers each word in documents as a feature. In this study, we use word frequency as the feature value to represent each sentence in a bag-of-words model.
- *Part-of-speech (POS)* is the grammatical property of words. It is a common feature for sentence classification (Chung, 2009; S. N. Kim et al., 2010; Lui, 2012; Naughton, Stokes, & Carthy, 2010). We use Jieba for Chinese POS tagging, which can extract 40 different part-of-speech tags. We attach the POS tag to each word to generate the POS features and use appearance frequency within each sentence as the feature value.
- *Latent Dirichlet Allocation (LDA)* is a probabilistic topic model (Blei, Ng, & Jordan, 2003), which can be used for topic clustering. LDA can give the probability that the sentence will have each topic. We apply LDA on the bag-of-words features to generate 40 groups, which is tuned using a sample of the dataset. We then consider the appearance of the sentences in the 40 groups as LDA features (S. Zhou & Li, 2018).
- *Doc2Vector* is the extension of Word2Vector for semantic analysis. Word2Vector employs two-layer neural networks to reconstruct linguistic contexts of words and produce a vector representation. In Word2Vector, words that share common contexts in the corpus are located closer in the vector space. Doc2Vector uses the same approach, generating vectors for sentences, paragraphs, or articles. In this study, we use genism (Radim & Sojka, 2010) and set the vector to 40 dimensions through tuning. Thus, each sentence is represented as a vector with 40 feature values.

Structural features capture the structural characteristics of the sentences in the paper. In our work, we choose three structural features: heading, sentence location, and sentence length.

- *Heading:* For structured content, section headings have important clues for section classification (S. N. Kim et al., 2010; Lui, 2012). However, in our study, the Chinese medical literature does not have well-structured section headings. Thus, we identify possible section headings based on heuristics. In particular, if a sentence is the only sentence of the paragraph with less than 6 words and no punctuation at the end, we define it as a section heading. For each section heading, we apply bag-of-words to generate features. The heading feature is associated with each sentence in the subsequent paragraphs until a new section heading is identified.

- *Sentence location:* Sentence location shows the position of the sentence in the paper and may indicate its role in the logic (Hirohata et al., 2008; Takahiko Ito et al., 2004; McKnight & Srinivasan, 2003; Naughton et al., 2010). For example, the introduction often appears early in the paper, while results and discussion are at the end. We use the normalized sentence position (counted by number of sentences) in the whole article as this feature.
- *Sentence length:* Previous studies (Hirohata et al., 2008; Lui, 2012; Naughton et al., 2010) used the number of words in a sentence as a feature for section identification. Because we are dealing with Chinese literature, we employ both the number of characters and the number of words in the sentence for this feature.

### *4.1.2 Textual Feature Selection and Sentence Dependency Modeling*

As shown in the literature, feature selection, which provides data free from irrelevant and redundant features (Rehman, Javed, & Babri, 2017), can improve performance and efficiency. For all the generated features, we conduct Information Gain (IG)-based feature selection due to its good performance. Information gain measures the richness of information in each feature for classification in terms of reducing the dataset's information entropy. Larger information gain suggests a feature's better classification ability.

After feature selection, we apply three machine learning models for the sentence classification task: Logistic Regression (LR), Support Vector Machine (SVM), and Conditional Random Fields (CRF). Logistic regression (Lui, 2012) and SVM (Takahiko Ito et al., 2004; McKnight & Srinivasan, 2003; Yamamoto & Takagi, 2005) represent the traditional models that ignore sentence interdependency. We implement both models using the Scikits-learn package in python (Pedregosa et al., 2011) and use them in a multi-class classification setup.

Conditional Random Fields (CRF) represent the models that makes use of sentence interdependency, which has been shown effective in literature. CRF is a type of discriminative undirected probabilistic graphical model. When inferring the distribution of random variables, CRF takes into consideration dependency between two random variables. In our study, sentences' classification labels have dependencies, since papers generally have an inherent logic (Chung, 2009; Hirohata et al., 2008; S. N. Kim et al., 2010). To capture such sentence interdependency, we include two sentences before and after each sentence in the CRF template as input to classify a focal sentence. The window size is decided through parameter tuning.

## 4.2 Deep Learning Approach (A SLSTM Model)

Deep learning is a different approach that does not need explicit feature generation. For the deep learning models, we take the word segmentation output as inputs to build the model, which is often needed to convert to word embedding in deep learning text mining. Since BLSTM is the state-of-the-art algorithm in deep learning text mining, it would be straightforward to employ this method to classify sentences. However, the model does not fully capture the characteristics of section identification. Thus, in this research, we propose a new deep learning model to address the problem.

In section identification, we need to consider not only the dependencies of words as in a regular sentence classification problem, but also the interdependencies of sentences, because text hierarchy is important in classification problems (Liu, 2009). There is also structural information that can be taken advantage of in this task. Based on these concerns, we propose a Structural Bidirectional Long Short-Term Memory (SLSTM) model as shown in Figure 3.

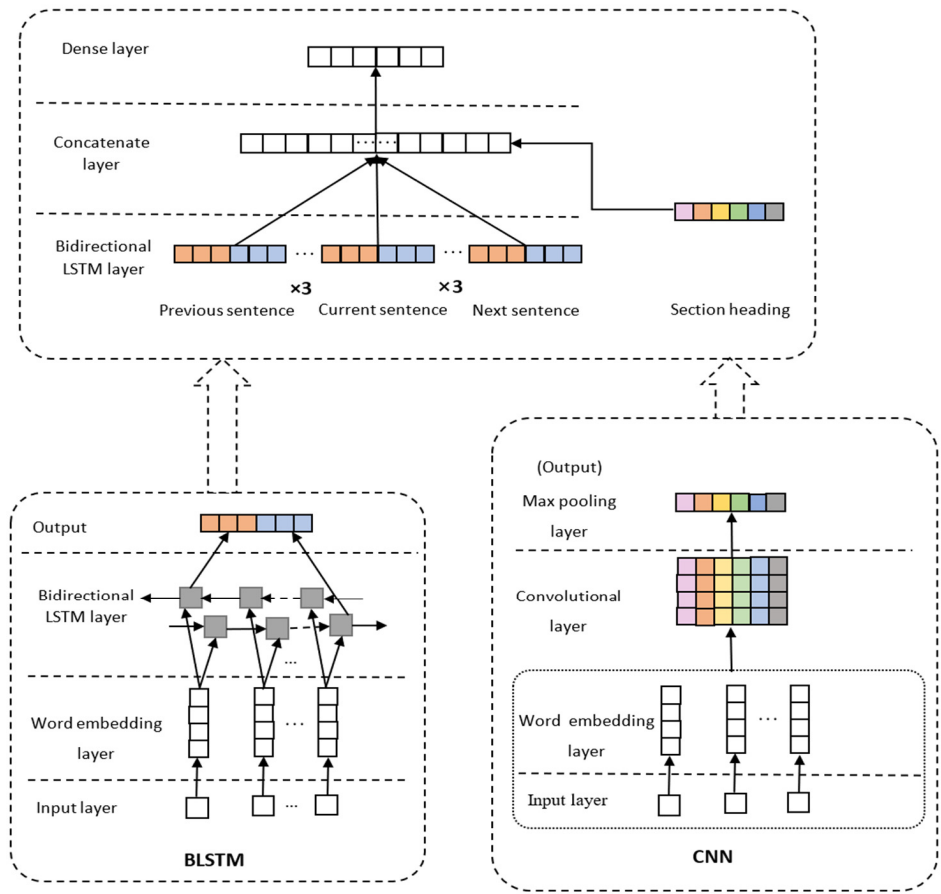

Figure 3. SLSTM Structure

The main idea of the SLSTM model is to use BLSTM to deal with word-level interdependencies and employ a fully connected neural network structure to model the sentence-level interdependencies and structure of paragraphs.

Our model has three main parts: the part to deal with word-level interdependencies within sentences, the part to deal with the interdependencies between neighboring sentences within the articles, the part to deal with the structure of articles. First, at an individual sentence level, we use pre-trained word embedding to preprocess the data for dimension reduction. Then we use a BLSTM layer to aggregate information into one output. In this part, the inputs are a series of words, while the outputs are aggregated information to represent the sentences. BLSTM can capture the sequence feature in the sentences.

Second, for each sentence, we make use of the nearby sentences' information as processed by BLSTM to aid the classification of the current sentence. In our study, we choose to include 3 sentences before and 3 sentences after the current sentence through tuning on a sample for the best performance. We use the same BLSTM layer to capture the information of total seven sentences. Then we employ a concatenate layer, which combines the current sentence with its related sentences.

Third, in order to make use of structural information, i.e., section headings, for section identification, the section headings of the current sentence are input to the concatenate layer as well. Section headings are usually short and lack complete sentence structure. Thus, we model them using CNN (Y. Kim, 2014), which can process short text better, while BLSTM may cause overfitting problems. Here, we use random word embedding (Gal & Ghahramani, 2015) for dimension reduction. Then, a convolutional layer and a max-pooling layer are used to abstract information in each section heading to one output. Finally, the concatenate layer is followed by a dense layer, which helps complete the classification task.

Our SLSTM model can be simplified to other models. If we remove the section heading part, we can make a model that only considers nearby sentences' interdependencies. We call it a Concatenate BLSTM (CLSTM) model. If we consider 0 nearby sentences in the model, the model is further reduced to a BLSTM model. We adopt Hierarchical Attention Networks (HAN) (Z. Yang et al., 2016) as the baseline since they are a well-known approach for similar problems. To consider the dependencies between sentences, we consider the current sentence, the 3 prior sentences, and the 3 next sentences as a short document and used classical HAN.

# 5 Evaluation

## 5.1 Dataset

To evaluate the performance of our proposed methods, we code a gold standard using a dataset provided by a Chinese physician. The dataset, focusing on the treatment of asthma with acupuncture from 1978 to 2008, was manually identified and downloaded from CNKI, China's largest literature database. We process the downloaded files, which are in pdf format, and convert them to text. We exclude files created from scanned documents that require OCR for conversion to avoid the noise caused by OCR. We remove footers, headers, and watermarks and get a set of clean text files covering the contents of the papers. To focus on the extraction of experiment-related content, we also remove positional papers or commentaries with no concrete experimental evidence. The final dataset contains 371 papers with 15,057 sentences.

We employ a research assistant to code the Chinese medical papers and label the sentences related to Subject, Method, and Result. A PhD student randomly sampled and checked the RA's coding. As the papers are written in modern Chinese and are generally easy to understand, the coding is of high quality. Eventually, our coded dataset has 1038 Subject sentences, 3122 Method sentences, and 1622 Result sentences. On average, each paper has 41 sentences, with 3 sentences related to Subject, 8 sentences related to Method, and 4 sentences related to Result.

## 5.2 Evaluation Metrics

We use precision (P), recall (R), and $F_1$-score to assess the section identification performance. As explained in section 3.2, since we aim to find useful information in Chinese medical literature, our focus is the identification of Subject, Method, or Result. For each label, precision is the ratio of sentences that are predicted correctly. Recall is the ratio of sentences in the gold standard that are correctly predicted. The $F_1$-score combines precision and recall as an overall assessment of the performance.

$$P = \frac{\#of\ correctly\ predicted\ sentences}{\#of\ predicted\ sentences\ for\ the\ label} \quad (1)$$

$$R = \frac{\#of\ correctly\ predicted\ sentences}{\#of\ sentences\ with\ the\ label\ in\ gold\ standard} \quad (2)$$

$$F_1 = \frac{2*P*R}{P+R} \quad (3)$$

### 5.3 Experimental Procedure

For the feature-based machine learning models, LR, SVM, and CRF, we tuned both the number of LDA classes for LDA features and the number of Doc2Vector dimensions for Doc2Vector features in small scale experiments. The results suggest the optimal value to be 40.

In addition, we did feature selection to select the most suitable combination of features to represent individual sentences. We use the information gain between the dependent variable and individual variable to judge the informativeness of each variable. Following that we set a threshold to filter out the low information gain variables. This is classic way for information gain-based feature selection (Y. Yang & Pedersen, 2004). In the experiments, we systematically experiment with different thresholds and compare the performance with selected features. Since our feature selection is at sentence level, it won't affect the modeling of sequence of sentences, such as in CRF. CRF considers the interdependency among sequential sentences through templates. The selected features for each sentence are the input which is combined with the templates to be modeled in CRF.

For the deep learning models, we compare our proposed SLSTM model with its simplifications, the CLSTM model and the BLSTM model. For the input, we set the word embedding dimension for individual sentences to be 200 for each word; the input feature length is 100 for the BLSTM model. In our dataset, the longest sentence has 97 words. The feature length is sufficient for the dataset. We set the input feature length for section headings to 5, because the length of section headings is usually less than 6. If the sentence is the first or last 2 sentences in the article, there are not sufficient related sentences. The inputs of the non-existing sentences are all zero vector. For the hyperparameter tuning, the network is trained by back-propagation in mini-batches, and the gradient-based optimization is performed using the Adam update rule. The initial learning rate is 0.001, dropout probability is 0.2, and batch size is 128. The number of nodes in each layer is 200, and the kernel size is 3 for the convolution layer.

After parameter tuning, we conduct 10-fold experiments at paper level and conduct statistical tests for performance evaluation.

# 6 Results

## 6.1 Feature-based Section Identification

### 6.1.1 Feature Selection

Figure 4 shows the performances with different thresholds in the CRF model. We also experiment with LR and SVM with the same set of IG threshold, which is in the Appendix Table A2. While for most of the performance measures on the three labels, the performance peaks if the IG threshold is set to 0.009. Under this setup, there are 170 features left, which include 68 BOW features, 87 POS features, and 15 heading features. As we can see, the performances of the models are generally stable when IG threshold is relaxed to 0.01. When reach this threshold, for each model (CRF, LR, and SVM) and each problem (Subject, Method, and Result), the performance (on precision, recall, and f-measure) are generally stable. It indicates that this parameter setup could be used as the starting point for future studies that want to tackle a similar problem on a different data set.

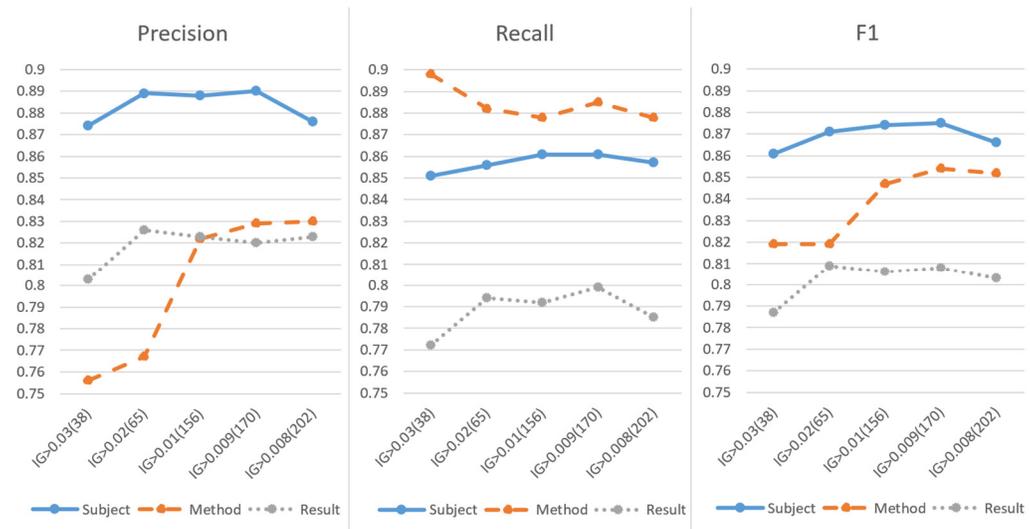

Figure 4. Feature Selection Performance in CRF

### 6.1.2 Classification Performance

Table 4 shows the performance of the selected feature set using LR, SVM, and CRF. (The results of two arbitrary labels, Pre and After, are reported in the Appendix Table A1). The CRF model achieves the highest performance on most of the dimensions. The $F_1$-scores of CRF with selected features are 87.5%, 85.4%, and 80.8%, respectively, for Subject, Method, and Result. We conduct a pair-wise t-test to evaluate the improvement in our results. CRF is significantly better than logistic regression and SVM in precision and $F_1$-score, while they do not have significant differences for recall on classifying Method and Result sentences. This result is consistent if we choose different threshold (or even remove the feature selection part) for these three models. The results show that interdependent information as captured by the CRF model is very important for section identification. It is necessary to consider the paper's structure and the dependencies between sentences for this task.

Table 4. Performance of LR, SVM, and CRF

| Model | Subject | | | Method | | | Result | | |
| --- | --- | --- | --- | --- | --- | --- | --- | --- | --- |
| | *P(%)* | *R(%)* | $F_1$*(%)* | *P(%)* | *R(%)* | $F_1$*(%)* | *P(%)* | *R(%)* | $F_1$*(%)* |
| LR | 81.7*** | **85.4** | 83.6*** | 70.0*** | **91.1** | 78.9*** | 76.6*** | **79.8** | 78.1** |
| SVM | 83.1*** | **83.9**** | 83.3*** | 70.4*** | **91.0** | 79.3*** | 77.0*** | **79.6** | 78.2** |
| CRF | **89.0** | **86.1** | **87.5** | **82.8** | **88.5** | **85.4** | **82.0** | **79.9** | **80.8** |

(p-value for comparison with CRF: ***<0.01; **<0.05; *<0.1. The bold values are not significantly different from the highest value in the column.)

**6.2 Deep Learning-based Section Identification**

Table 5 shows the performance of our proposed SLSTM model. For comparison, we also include BLSTM, HAN, CLSTM, and CRF as baselines. Our SLSTM model has the highest performance in all dimensions. We conduct a pair-wise t-test to compare the different models. First, as compared with the three deep learning baselines, BLSTM, HAN, and CLSTM, our proposed model is significantly better (except that SLSTM and CLSTM have insignificant differences in precision to classify Subject, SLSTM and HAN have insignificant differences in precision to classify Method). The advantage of SLSTM over HAN and CLSTM shows that section headings do help in section identification. The advantage of our model over BLSTM, the state-of-the-art deep learning algorithm, further shows the importance of considering the dependencies between sentences in this task.

Second, we compare SLSTM and the best feature-based method, CRF. Our SLSTM model performs better than CRF in all dimensions except in precision of Subject classification. Moreover, there is a big performance improvement numerically. For example, the $F_1$-scores increased 1.49%, 3.71%, and 3.16%, respectively, for Subject, Method, and Result. The performances show the good potential of deep learning models for the section identification task.

The classic BLSTM model is not as good as the CRF model, which shows that significant modeling efforts are still needed in the world of deep learning.

Table 5. Performance of Deep Learning Models and CRF

| Model | Subject | | | Method | | | Result | | |
| --- | --- | --- | --- | --- | --- | --- | --- | --- | --- |
| | *P(%)* | *R(%)* | *$F_1$(%)* | *P(%)* | *R(%)* | *$F_1$(%)* | *P(%)* | *R(%)* | *$F_1$(%)* |
| CRF | **89.0** | 86.1** | 87.5** | 82.8*** | 88.5* | 85.4*** | 82.0*** | 79.9* | 80.8*** |
| BLSTM | 80.9*** | 81.1*** | 80.8*** | 77.4*** | 73.6*** | 74.9*** | 76.1*** | 59.7*** | 65.7*** |
| HAN | 80.6*** | 80.5*** | 80.1*** | 86.8 | 85.4*** | 85.9*** | 77.8*** | 74.6*** | 75.7*** |
| CLSTM | **87.3** | 88.1** | 87.3** | 87.0* | 86.3*** | 86.3*** | 82.5** | 78.7** | 80.4*** |
| SLSTM | **89.0** | **89.4** | **89.0** | **88.2** | **90.0** | **89.1** | **86.0** | **81.9** | **84.0** |

(p-value for comparison with SLSTM: ***<0.01; **<0.05; *<0.1. The bold values are not significantly different from the highest value in the column.)

One may concern that the SLSTM has a relatively simple structure. To investigate the effectiveness of our model, we also adopted many different models and compare their results. In addition to the hieratical attention network model, we also consider double layer BLSTM framework and incorporated section heading identification into the deep learning framework. However, SLSTM model beat all these efforts, showing that our purposed model has captured the useful characteristics of this problem. While we believe there is always a room to improve the model structure of a problem, we leave such explorations to future research.

One may also concern that our model does not explicitly employ the medical knowledge. To address this concern, we conducted experiments by considering the medical terms (disease, medicine, and acupuncture points) into word embedding or as additional input to the concatenate layer. However, these more complicate models' performances are not improved much from our original model. This finding shows that the deep learning framework, can capture the information needed for the section identification very well.

To further illustrate the effectiveness of our SLSTM model in capturing the essence of medical information, we cluster words using their word embedding vector. After some tweaking we set the number of clusters to 5. In Figure 5 we visualize the clusters using their two major principal components. We also show the semantics of the clusters in Table 6. As we can see, the differentiation of the five clusters are clear. Cluster 4 is very distinct from other clusters, which is the terms for diseases and biological terms (that are shared between Chinese medicine and

Western medicine). Clusters 1 & 2 are also clearly separated from Clusters 3 & 5, where Clusters 1 and 2 are about subjects and experiments, and Clusters 3 & 5 are about acupuncture points and Chinese medicines, which are unique to Chinese traditional medicine. Clearly, the deep learning word vectors can, to a large extent, differentiate the semantics of different terms.

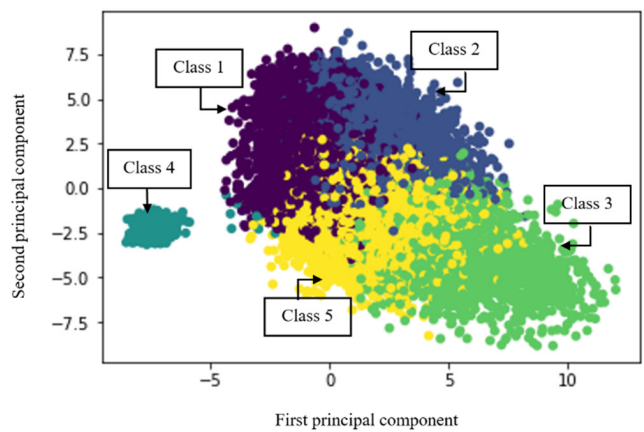

Figure 5. Cluster Visualization

Table 6. Semantics of Clusters

| Label | Some high frequency words (Translated from Chinese) |
|---|---|
| 1- subject, time and others | year, woman, man, hour, every time, author... |
| 2-experiment and performance | method, standard, average, significance, frequency, statistics, level, regulation, reduction, ineffective, improvement, mitigation, effective... |
| 3- acupuncture point and related treatment | Yu point, Pei point, Dazhui point, Tiantu point, Jing point… |
| 4-disease, symptoms and biological terms | Asthma, bronchus, symptoms, diagnosis, cough, morbidity, illness, cough, cells, granulocytes, lymphocytes, white blood cells… |
| 5-traditional Chinese medicine and related treatment | ginger juice, ephedra, application, traditional Chinese medicine, application, press... |

## 6.3 Effect of Section Identification on Literature Analysis

To illustrate the effect of applying our algorithm in applications, we apply the SLSTM model trained with the 371 gold standard data on 6169 Chinese medicine papers on asthma published from 2010 to 2018 to filter the irrelevant sentences. Then, we apply a dictionary-based entity recognition algorithm to extract medical keywords, including disease, medicine, and acupoints (i.e., acupuncture points) from the filtered and unfiltered data.

Table 7. Entity Extraction on Unfiltered and Filtered Data

|  | *Sentences* | *Medical Keywords* | *Disease* | *Medicine* | *Acupoint* |
|---|---|---|---|---|---|
| Unfiltered | 355,598 | 799 | 189 | 388 | 222 |
| Filtered | 68,305 | 533 | 92 | 321 | 120 |
| % | 19.21% | 66.70% | 48.68% | 82.73% | 54.05% |

Table 7 shows the summary statistics of the filtered and unfiltered sentences and keywords. After filtering, our model kept 19.21% of all 355,598 sentences and 66.70% of keywords. Among these extracted keywords, 82.73% of medicine words are kept. In a medical paper, medicines often appear in the treatment section. This indicates that our model can remove irrelevant discussions and keep the main findings for analysis.

To further illustrate the effect of section identification on literature analysis, we examine the co-occurrence relationship of keywords within papers. In co-occurrence analysis, it is a common practice to examine the high frequency relations, which removes the unimportant relations and noise relations that appeared accidentally in the dataset. In Table 8, we tweak the threshold to be 2, 5, 10, and 20, respectively, and examine how our section identification affected the identified co-occurrence relations under different thresholds.

We notice that by increasing the threshold, the identified relations are reduced in both filtered and unfiltered data, which leaves a set of increasingly trustworthy relations for analysis. Meanwhile, the overlap of the relations of the filtered and unfiltered data actually increases with threshold increases. For example, on the entire dataset, the filtered data contains only 30% of the relations from the unfiltered dataset. But if we only consider the relationships that appear more than 20 times, the overlap of filtered and unfiltered data is 98.4%, which shows that our filtered data kept the most important relations in the literature. Moreover, if researchers want to study relationships that appear less frequently in the literature, using the filtered data can significantly reduce their risk of getting noise.

Table 8. Co-occurrence Relations on Unfiltered and Filtered Data

| Co-occurrence Relationship | | Total | Disease-Disease | Medicine-Medicine | Acupoint-Acupoint | Disease-Medicine | Disease-Acupoint | Medicine-Acupoint |
|---|---|---|---|---|---|---|---|---|
| Frequency>0 | Unfiltered | 66,469 | 4,679 | 25,491 | 6,906 | 12,182 | 4,425 | 12,786 |
| | Filtered | 20,905 | 381 | 12,701 | 1,447 | 2,723 | 630 | 3,023 |
| | % | 31.45 | 8.14 | 49.8 | 20.95 | 22.35 | 14.23 | 23.64 |
| Frequency>2 | Unfiltered | 34,885 | 1,109 | 16,820 | 3,422 | 5,344 | 1,570 | 6,620 |
| | Filtered | 18,558 | 328 | 11,369 | 1,287 | 2,350 | 558 | 2,666 |
| | % | 53.20 | 29.58 | 67.59 | 37.61 | 43.97 | 35.54 | 40.27 |
| Frequency>5 | Unfiltered | 17,550 | 341 | 9,730 | 1,389 | 2,453 | 636 | 3,001 |
| | Filtered | 13,973 | 236 | 8,551 | 980 | 1,770 | 428 | 2,008 |
| | % | 79.62 | 69.21 | 87.88 | 70.55 | 72.16 | 67.30 | 66.91 |
| Frequency>10 | Unfiltered | 10,705 | 189 | 6,218 | 752 | 1,503 | 366 | 1,677 |
| | Filtered | 9,936 | 166 | 6,045 | 666 | 1,322 | 304 | 1,433 |
| | % | 92.80 | 87.80 | 97.22 | 88.57 | 87.96 | 83.06 | 85.45 |
| Frequency>20 | Unfiltered | 6,317 | 107 | 3,754 | 434 | 931 | 209 | 882 |
| | Filtered | 6,216 | 105 | 3,744 | 421 | 904 | 204 | 838 |
| | % | 98.40 | 98.13 | 99.73 | 97.00 | 97.10 | 97.61 | 95.01 |

To have a more detailed view of the extracted relations, we visualize the disease co-occurrence network using Ucinet (Borgatti, Everett, & Freeman, 2002). Figure 6 visualizes the network using the filtered data with co-occurrence frequency higher than 10 and chooses the same entities in the unfiltered data. While the two networks have similar nodes and structures, the filtered network has significantly fewer relations. Closer inspection of the relations in the network shows that the filtered dataset makes better sense. For example, the connections between cold and asthma, cough, and allergic rhinitis are well known. But the relationships between cold and tuberculosis and lung cancer (exists only in the unfiltered network) are due to lack of sufficient analysis support for Chinese medical research.

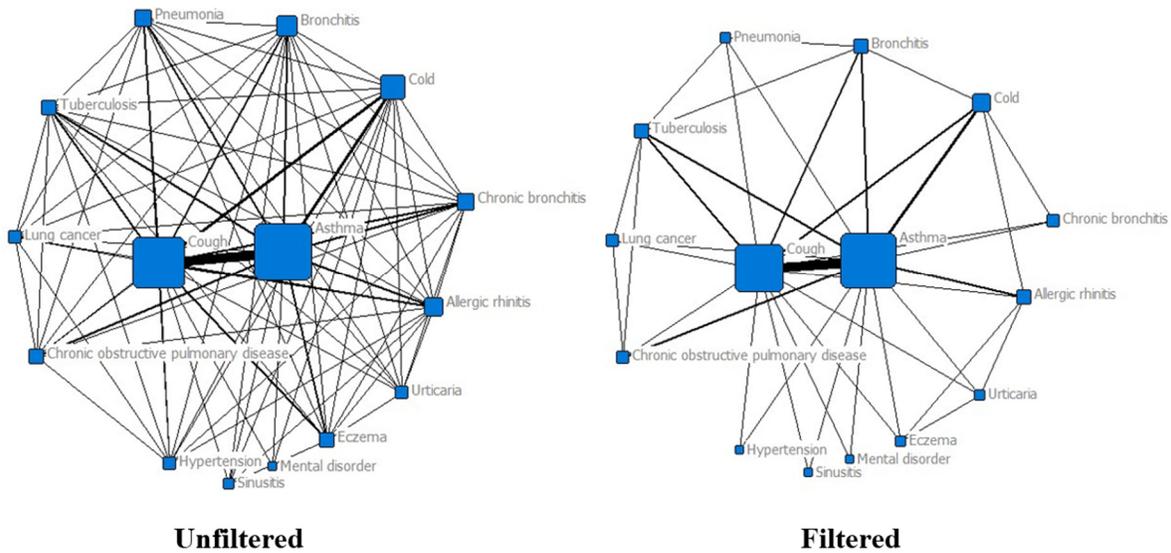

Figure 6. Disease Co-occurrence Networks (Translated from Chinese)

In general, the case study shows that our section identification model can help keep the important sentences in literature for analysis while removing less relevant sentences to save researchers' effort.

## 7 Conclusion

In this research, we developed new models to identify sections in Chinese medical literature to support analysis tasks in knowledge management. We take both a feature engineering approach and a deep learning approach to tackle the problem. With carefully developed features, we found that CRF outperforms other classic machine learning methods, benefitting from capturing dependencies between sentences. Based on this observation, we propose a SLSTM model, which considers the dependencies between sentences in a deep learning framework. Experiments show that the SLSTM model can achieve close to 90% performance in precision, recall, and $F_1$-measures and significantly outperforms other deep learning methods and the CRF method. Experiments show that our model can extract the most important entities and relations from Chinese medical literature for future analysis. It also shows the potential of employing deep learning for the section identification problem. It is noted that classic machine learning models still have advantages unless the deep learning models are carefully built.

This study offers important theoretical and practical implications for section identification and knowledge management. From a theoretical perspective, we find that it is worthwhile to model the structure of articles and the interdependencies between sentences to tackle the section identification problem, whether using a deep learning approach or a feature engineering approach. This extends the findings in previous studies which are mostly under a feature engineering framework. Moreover, we find that generic deep learning models do not always perform better than classic machine learning models. Problem-specific knowledge needs to be carefully modeled to address text mining problems. (In our case, sentence interdependency is effective problem-specific knowledge.) Deep learning does not guarantee good performance.

From a practical perspective, we provide practical guidelines on how to address the section identification problem in/for Chinese medical papers, which is an important and less studied application area. In the study, we found the effective features for the feature engineering approach and propose an effective SLSTM model for the deep learning approach. Both methods can reduce human effort and alleviate the information overload problem in information processing. The algorithms can help scholars make the most use of literature and design more effective Chinese medical literature analysis systems, since relevant sentence extraction in the medical domain is necessary to save clinicians' time (Jonnalagadda et al., 2012). Moreover, our proposed SLSTM model has the potential to address other sentence classification problems. As we know, many problems in knowledge management are related to sentence understanding and classification. Our proposed model thus has the potential to benefit information science research and practitioners in general.

This study offers important implications for section identification and knowledge management. First, we extend the section identification issue to traditional Chinese medical papers, which is an important and less studied application area. While information extraction can help scholars make the most use of digital literature, our identified effective features and proposed effective models would significantly help future text mining studies on Chinese medical literature. Second, from a methodological perspective, we found that the dependencies between sentences and the structure of articles are important aspects that need to be captured to successfully tackle the section identification problem, such as in our SLSTM and the CRF model. By extending previous studies that are mostly on medical abstracts, our findings on the full text of documents provide stronger evidence that such information would be necessary for section identification in other similar tasks. Third, we propose a SLSTM with good performance. It has potential to be used to address other sentence classification problems.

The study also has its limitations that can be improved in future research. First, our gold standard dataset only has 371 papers. Enlarging the dataset may benefit deep learning models and further the power of deep learning. Second, our model only employs textual information. Other information in literature, such as figures, tables, etc., can help address the problem. Third, this study takes a sequential approach to model inter-sentence dependency. There is a room to improve the model structure, such as to model longer sentences and more complicate sentence interdependencies. Such efforts may benefit the performance. Third, in this study, we do not fully exploit Chinese linguistic characteristics, such as synonyms (for example, one Chinese medicine may have multiple names), which has been applied in English text analysis (Plaza, Stevenson, & Díaz, 2012). It is possible to model features in the task, which can be investigated in future research.

## Acknowledgements

The research was partially supported by Guangdong Science and Technology Project 2014A020221090, CityU SRG 7005195, the City University of Hong Kong Shenzhen Research Institute, and the Digital Innovation Lab at City University of Hong Kong.

## References

Agibetov, A., Blagec, K., Xu, H., & Samwald, M. (2018). Fast and scalable neural embedding models for biomedical sentence classification. *BMCBioinformatics*, *19*(1), 1–9.
Angrosh, M. a., Cranefield, S., & Stanger, N. (2010). Context identification of sentences in related work sections using a conditional random field: towards intelligent digital libraries. In *Proceedings of the 10th annual joint conference on Digital libraries. ACM* (pp. 293–302). https://doi.org/10.1145/1816123.1816168
Blei, D. M., Ng, A. Y., & Jordan, M. I. (2003). Latent Dirichlet Allocation. *Journal of Machine Learning Research*, *3*, 993–1022. https://doi.org/10.1162/jmlr.2003.3.4-5.993
Borgatti, S. P., Everett, M. G., & Freeman, L. C. (2002). UCINET for Windows : Software for social network analysis. *Analytic Technologies*. Harvard, MA.
Chung, G. Y. (2009). Sentence retrieval for abstracts of randomized controlled trials. *BMC Medical Informatics and Decision Making*, *9*(1), 1–13. https://doi.org/10.1186/1472-6947-9-10
Chung, G. Y., & Coiera, E. (2007). A study of structured clinical abstracts and the semantic classification of sentences. *Proceedings of the Workshop on BioNLP 2007 Biological, Translational, and Clinical Language Processing - BioNLP '07*, (June), 121. https://doi.org/10.3115/1572392.1572415
Dernoncourt, F., L, J. Y., & Szolovits, P. (2017). Neural Networks for Joint Sentence Classification in Medical Paper Abstracts. In *Proceedings of the 15th Conference of the European Chapter of the Association for Computational Linguistics* (pp. 694–700).

El-Gayar, O., & Timsina, P. (2014). Opportunities for Business Intelligence and Big Data Analytics in Evidence Based Medicine. *2014 47th Hawaii International Conference on System Sciences*, 749–757. https://doi.org/10.1109/HICSS.2014.100

Esuli, A., Marcheggiani, D., & Sebastiani, F. (2013). An enhanced CRFs-based system for information extraction from radiology reports q. *Journal of Biomedical Informatics*, *46*(3), 425–435. https://doi.org/10.1016/j.jbi.2013.01.006

Gal, Y., & Ghahramani, Z. (2015). *A Theoretically Grounded Application of Dropout in Recurrent Neural Networks*. https://doi.org/10.1201/9781420049176

Hachey, B., & Grover, C. (2005). Sequence modelling for sentence classification in a legal summarisation system. In *Proceedings of the 2005 ACM symposium on Applied computing* (pp. 292–296). https://doi.org/10.1145/1066677.1066746

Hassanzadeh, H., Groza, T., & Hunter, J. (2014). Identifying scientific artefacts in biomedical literature: The Evidence Based Medicine use case. *Journal of Biomedical Informatics*, *49*, 159–170. https://doi.org/10.1016/j.jbi.2014.02.006

Hirohata, K., Okazaki, N., Ananiadou, S., & Ishizuka, M. (2008). Identifying Sections in Scientific Abstracts using Conditional Random Fields. In *Proceedings of the Third International Joint Conference on Natural Language Processing: Volume-I.* (pp. 381–388).

Ito, T., Shimbo, M. A. S., & Amasaki, T. A. Y. (2004). Semi-supervised sentence classification for MEDLINE documents. *METHODS*, *138*(25), 141–146.

Ito, T., Shimbo, M., Yamasaki, T., & Matsumoto, Y. (2004). Semi-supervised sentence classification for MEDLINE documents. *METHODS*, *138*(25), 141–146.

Jonnalagadda, S. R., Fiol, G. Del, Medlin, R., Weir, C., Fiszman, M., Mostafa, J., & Liu, H. (2012). Automatically extracting sentences from Medline citations to support clinicians' information needs. *Journal of the American Medical Informatics Association*, *20*(5), 995–1000. https://doi.org/10.1136/amiajnl-2012-001347

Kim, S. N., Martinez, D., & Cavedon, L. (2010). Automatic Classification of Sentences for Evidence Based Medicine. In *Proceedings of the ACM fourth international workshop on Data and text mining in biomedical informatics* (pp. 13–22). https://doi.org/10.1145/1871871.1871876

Kim, Y. (2014). Convolutional Neural Networks for Sentence Classification. In *Proceedings of the 2014 Conference on Empirical Methods in Natural Language Processing (EMNLP)* (pp. 1746–1751). https://doi.org/10.3115/v1/D14-1181

Lai, S., Xu, L., Liu, K., & Zhao, J. (2015). Recurrent Convolutional Neural Networks for Text Classification. *Twenty-Ninth AAAI Conference on Artificial Intelligence*, 2267–2273.

Lin, J., Karakos, D., Demner-fushman, D., & Khudanpur, S. (2006). Generative Content Models for Structural Analysis of Medical Abstracts. *BioNLP Workshop on Linking Natural Language Processing and Biology: Towards Deeper Biological Literature Analysis*, (June), 65–72. https://doi.org/10.3115/1567619.1567631

Liu, R. (2009). Context Recognition for Hierarchical Text Classification. *Journal of the Association for Information Science and Technology*, *60*(January), 803–813. https://doi.org/10.1002/asi

Lui, M. (2012). Feature Stacking for Sentence Classification in Evidence-Based Medicine. In *Proceedings of the Australasian Language Technology Association Workshop 2012* (pp. 134–138). Retrieved from http://aclweb.org/anthology/U12-1019

Malmasi, Hassanzadeh, H., & Dras, M. (2015). Clinical Information Extraction Using Word Representations. *Proceedings of the Australasian Language Technology Association*

*Workshop 2015*, 66–74. Retrieved from http://aclweb.org/anthology/U15-1008

McKnight, L., & Srinivasan, P. (2003). Categorization of sentence types in medical abstracts. In *AMIA Annual Symposium Proceedings. American Medical Informatics Association* (pp. 440–444). https://doi.org/D030003164 [pii]

Nam, S., Jeong, S., Kim, S. K., Kim, H. G., Ngo, V., & Zong, N. (2016). Structuralizing biomedical abstracts with discriminative linguistic features. *Computers in Biology and Medicine*, *79*(May), 276–285. https://doi.org/10.1016/j.compbiomed.2016.10.026

Naughton, M., Stokes, N., & Carthy, J. (2010). Sentence-level event classification in unstructured texts. *Information Retrieval*, *13*(2), 132–156. https://doi.org/10.1007/s10791-009-9113-0

Nguyen, H. T., & Nguyen, M. Le. (2018). Multilingual opinion mining on YouTube – A convolutional N-gram BiLSTM word embedding. *Information Processing & Management*, *54*(3), 451–462. https://doi.org/10.1016/j.ipm.2018.02.001

Pedregosa, F., Varoquaux, G., Gramfort, A., Michel, V., Thirion, B., Grisel, O., … Duchesnay, E. (2011). Scikit-learn : Machine Learning in Python. *Journal OfMachine Learning Research*, *12*, 2825–2830.

Plaza, L., Stevenson, M., & Díaz, A. (2012). Resolving ambiguity in biomedical text to improve summarization. *Information Processing & Management*, *48*(4), 755–766. https://doi.org/10.1016/j.ipm.2011.09.005

Radim, R., & Sojka, P. (2010). Software Framework for Topic Modelling with Large Corpora. In *Proceedings of the LREC 2010 Workshop on New Challenges for NLP Frameworks*.

Rehman, A., Javed, K., & Babri, H. A. (2017). Feature selection based on a normalized difference measure for text classification. *Information Processing & Management*, *53*, 473–489. https://doi.org/10.1016/j.ipm.2016.12.004

Ruch, P., Boyer, C., Chichester, C., Tbahriti, I., Geissbühler, A., Fabry, P., … Veuthey, A. L. (2006). Using argumentation to extract key sentences from biomedical abstracts. *International Journal of Medical Informatics*, *76*(2–3), 195–200. https://doi.org/10.1016/j.ijmedinf.2006.05.002

Shi, Y., & Bei, Y. (2019). HClaimE : A tool for identifying health claims in health news. *Information Processing & Management*, *56*(4), 1220–1233. https://doi.org/10.1016/j.ipm.2019.03.001

Shimbo, M., Yamasaki, T., & Matsumoto, Y. (2003). Using sectioning information for text retrieval: a case study with the Medline abstracts. *Proc. Second International Workshop on Active Mining (AM'03)*, 32–41.

Wu, J.-C., Chang, Y.-C., Liou, H.-C., & Chang, J. S. (2006). Computational Analysis of Move Structures in Academic Abstracts. *Proceedings of the COLING/ACL on Interactive Presentation Sessions*, (July), 41–44. https://doi.org/10.3115/1225403.1225414

Xu, R., Supekar, K., Huang, Y., Das, A., & Garber, A. (2006). Combining text classification and Hidden Markov Modeling techniques for categorizing sentences in randomized clinical trial abstracts. *AMIA ... Annual Symposium Proceedings / AMIA Symposium. AMIA Symposium*, 824–828. Retrieved from http://www.pubmedcentral.nih.gov/articlerender.fcgi?artid=1839538&tool=pmcentrez&rendertype=abstract

Yamamoto, Y., & Takagi, T. (2005). A sentence classification system for multi biomedical literature summarization. In *Proceedings of the 21st International Conference on Data Engineering* (pp. 1163–1168). https://doi.org/10.1109/ICDE.2005.170

Yang, Y., & Pedersen, J. O. (2004). A comparative study of feature selection and multiclass classification methods for tissue classification based on gene expression. *Bioinformatics*, *20*(15), 2429–2437. https://doi.org/10.1093/bioinformatics/bth267

Yang, Z., Yang, D., Dyer, C., He, X., Smola, A., & Hovy, E. (2016). Hierarchical Attention Networks for Document Classification. *Proceedings of the 2016 Conference of the North American Chapter of the Association for Computational Linguistics: Human Language Technologies*, 1480–1489. https://doi.org/10.18653/v1/N16-1174

Yin, W., & Hinrich Schütze. (2015). Multichannel Variable-Size Convolution for Sentence Classification. In *Proceedings of the 19th Conference on Computational Language Learning* (pp. 204–214).

Zhang, H., Boons, F., & Batista-navarro, R. (2019). Whose story is it anyway ? Automatic extraction of accounts from news articles. *Information Processing & Management*, (February), 1–12. https://doi.org/10.1016/j.ipm.2019.02.012

Zhang, X., Zhao, J., & LeCun, Y. (2015). Character-level Convolutional Networks for Text Classification. *Advances in Neural Information Processing Systems*, 3057–3061. https://doi.org/10.1063/1.4906785

Zhou, L., Ticrea, M., & Hovy, E. (2005). Multi-document Biography Summarization. In *Proceedings of the Empirical Methods in Natural Langauge Processing* (pp. 434–441). Retrieved from http://arxiv.org/abs/cs/0501078

Zhou, P., Shi, W., Tian, J., Qi, Z., Li, B., Hao, H., & Xu, B. (2016). Attention-Based Bidirectional Long Short-Term Memory Networks for Relation Classification. *Proceedings of the 54th Annual Meeting of the Association for Computational Linguistics (Volume 2: Short Papers)*, 207–212. https://doi.org/10.18653/v1/P16-2034

Zhou, S., & Li, X. (2018). Section Identification to Improve Information Extraction from Chinese Medical Literature. In *International Conference on Smart Health.* (pp. 342–350). Springer, Cham. https://doi.org/10.1007/978-3-030-03649-2_34

Zhou, X., Peng, Y., & Liu, B. (2010). Text mining for traditional Chinese medical knowledge discovery: A survey. *Journal of Biomedical Informatics*, *43*(4), 650–660. https://doi.org/10.1016/j.jbi.2010.01.002

# Appendix

A1. The feature selection results of LR and SVM

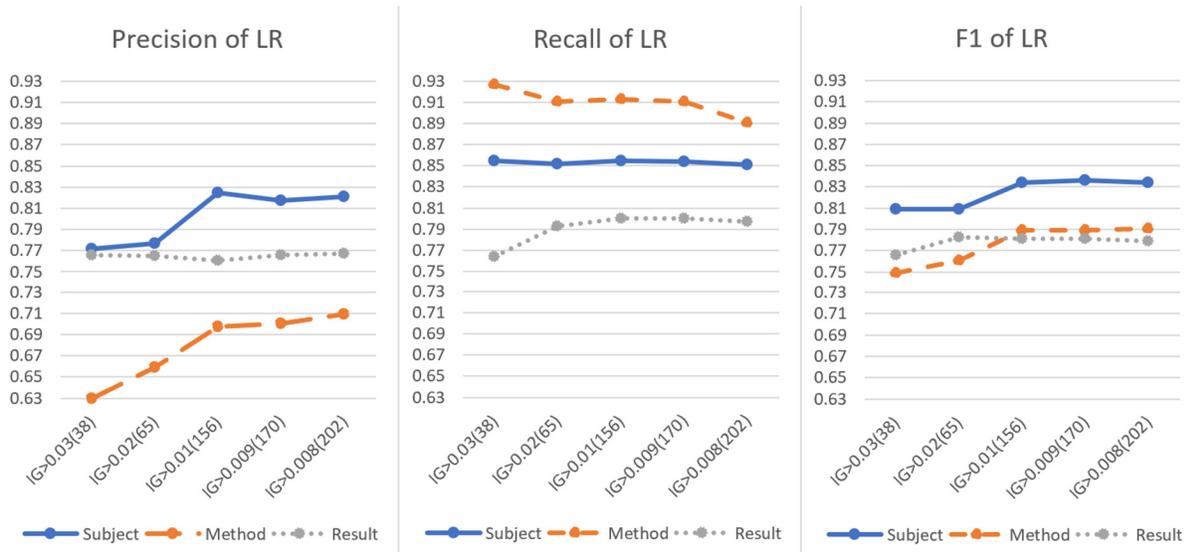

Figure A1. Feature Selection Performance in LR

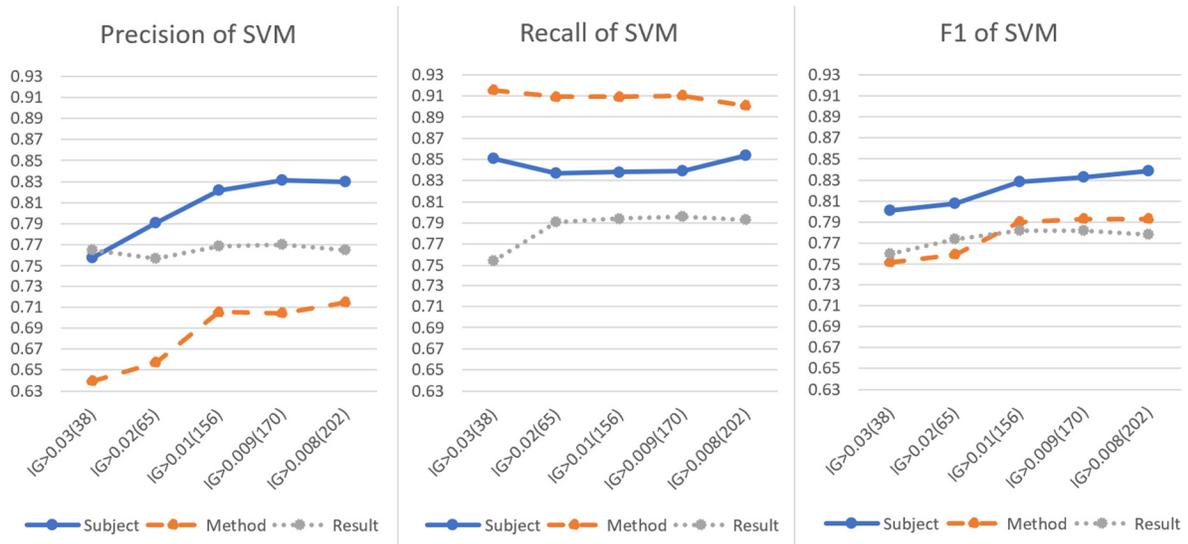

Figure A2. Feature Selection Performance in SVM

A2. The performance of different model on Pre and After labels

Table A1. Performance of the models on Pre and After labels

| Model | *Pre* | | | *After* | | |
|---|---|---|---|---|---|---|
| | *P(%)* | *R(%)* | *F₁(%)* | *P(%)* | *R(%)* | *F₁(%)* |
| LR | 0.786*** | 0.718*** | 0.749*** | **0.957** | 0.892*** | 0.924*** |
| SVM | 0.784*** | 0.739*** | 0.758*** | 0.953 | 0.881*** | 0.916*** |
| CRF | 0.896* | 0.882* | 0.887 | 0.947* | 0.942 | 0.945** |
| BLSTM | 0.770*** | 0.665*** | 0.706*** | 0.825*** | 0.932*** | 0.828*** |
| HAN | 0.899*** | 0.88** | 0.888*** | 0.893* | 0.927 | 0.91* |
| CLSTM | 0.926 | 0.867*** | 0.899*** | 0.924*** | 0.955** | 0.94*** |
| **SLSTM** | **0.944** | **0.898** | **0.919** | 0.955 | **0.965** | **0.957** |

(p-value for comparison with SLSTM: ***<0.01; **<0.05; *<0.1. The bold values are the largest of the column, which are not significantly different from the performance of SLSTM.)